\documentclass{article}

\usepackage[final]{neurips_2024}

\usepackage[utf8]{inputenc} % allow utf-8 input
\usepackage[T1]{fontenc}    % use 8-bit T1 fonts
\usepackage{hyperref}       % hyperlinks
\usepackage{url}            % simple URL typesetting
\usepackage{booktabs}       % professional-quality tables
\usepackage{amsfonts}       % blackboard math symbols
\usepackage{nicefrac}       % compact symbols for 1/2, etc.
\usepackage{microtype}      % microtypography
\usepackage[dvipsnames]{xcolor}

\usepackage{graphicx}
\usepackage[ruled,vlined]{algorithm2e}

\usepackage{inconsolata}
\usepackage{multirow}
\usepackage{tabularx}
\usepackage{colortbl}
\usepackage{pifont}
\usepackage{multicol}
\usepackage{booktabs}
\usepackage{makecell}

\usepackage{amssymb}
\usepackage{adjustbox}
\usepackage{enumerate}
\usepackage{amsmath}

\usepackage{placeins}

\definecolor{myred}{HTML}{C00000}
\definecolor{myorange}{HTML}{ED7D31}
\definecolor{mygreen}{HTML}{548235}
\definecolor{myblue}{HTML}{00B0F0}
\definecolor{mypurple}{HTML}{7030A0}

\definecolor{mydarkgreen}{rgb}{0.0, 0.5, 0.0}

\usepackage{marvosym}
\usepackage{utfsym}
\usepackage{xspace}
\usepackage{pifont}
\usepackage{arydshln}
\usepackage[T1]{fontenc}
\definecolor{LGray}{gray}{0.97}

\usepackage{sidecap}
\usepackage{booktabs}
\renewcommand{\arraystretch}{1}
\usepackage{afterpage}

\newcommand{\ie}{\textit{i.e.}}
\newcommand{\eg}{\textit{e.g.}}

\title{ZOPP: A Framework of Zero-shot Offboard Panoptic Perception for Autonomous Driving}

\author{
  Tao Ma$^{1,2}$\thanks{Equally contributed to the work.} \And Hongbin Zhou$^{2 *}$ \And Qiusheng Huang$^{2*}$ \And Xuemeng Yang$^2$ \And Jianfei Guo$^2$ \And Bo Zhang$^2$ \And Min Dou$^2$ \And Yu Qiao$^2$ \And Botian Shi$^2$ \And Hongsheng Li$^{1,3}$ \\
  [2mm]
  $^1$Multimedia Laboratory, The Chinese University of Hong Kong \\ $^2$Shanghai Artificial Intelligence Laboratory \ $^3$CPII \\
  [2mm]
  \texttt{taoma@link.cuhk.edu.hk, hsli@ee.cuhk.edu.hk} \\
  \texttt{\{zhouhongbin, huangqiusheng, yangxuemeng, shibotian\}@pjlab.org.cn}
}

\begin{document}

\maketitle

\begin{abstract}
  Offboard perception aims to automatically generate high-quality 3D labels for autonomous driving (AD) scenes.
  Existing offboard methods focus on 3D object detection with closed-set taxonomy and fail to match human-level recognition capability on the rapidly evolving perception tasks.
  Due to heavy reliance on human labels and the prevalence of data imbalance and sparsity, a unified framework for offboard auto-labeling various elements in AD scenes that meets the distinct needs of perception tasks is not being fully explored.
  In this paper, we propose a novel multi-modal Zero-shot Offboard Panoptic Perception (ZOPP) framework for autonomous driving scenes.
  ZOPP integrates the powerful zero-shot recognition capabilities of vision foundation models and 3D representations derived from point clouds.
  To the best of our knowledge, ZOPP represents a pioneering effort in the domain of multi-modal panoptic perception and auto labeling for autonomous driving scenes.
  We conduct comprehensive empirical studies and evaluations on Waymo open dataset to validate the proposed ZOPP on various perception tasks.
  To further explore the usability and extensibility of our proposed ZOPP, we also conduct experiments in downstream applications.
  The results further demonstrate the great potential of our ZOPP for real-world scenarios. The source code will be released at \url{https://github.com/PJLab-ADG/ZOPP}.
\end{abstract}

\section{Introduction}
\label{sec:intro}
Comprehensive perception and understanding of 3D scenes are important for autonomous driving (AD).
We have witnessed the evolution of machine perception at different levels within a short period: from single-modal~\cite{voxelnet,second,pointpillars,pointrcnn,centerpoint,pvrcnn,parta2,li2022bevformer,ma2024velovox,bai2023range,chen2022mppnet,chen2023trajectoryformer} to multi-modal inputs~\cite{logonet,uniseg,cai2021semantic,ma2021crlf,ma2021pe,wen2023dilu}, from limited categories to open set~\cite{liu2023grounding,li2023-openseg,lu2023smkm,pointclip,mluc,ov-detr,fm-ov3d}, from 3D box to 3D occupancy~\cite{li2023voxformer,sima2023_occnet,tian2023occ3d,huang2023tri,miao2023occdepth,yuan2024reg,yan2023spot}, and from low-level detection to high-level understanding.
Though remarkable, to train a model for different AD perception tasks, huge amounts of high-quality data and labels are still required, which is time-consuming and expensive.
Therefore, it is essential to come up with an efficient solution to cope with such rapid changes.

\begin{table*}[ht]
    \renewcommand\arraystretch{1.2}
    \centering
    \caption{Comparisons of recent onboard and offboard perception models.
    \textit{Seg.}, \textit{Det.}, \textit{Occ.} represent 3D segmentation, 3D object detection, occupancy prediction, respectively.
    \textit{HLF} means training in a human-label-free manner.
    \textit{Grounding} highlights models that can respond with text prompts.
    \textit{Zero.} stands for the zero-shot capability for unseen classes.
    }
    \label{tab:methods_comparison}
    \begin{tabular}{lccccccccc}
    \Xhline{0.75pt}
    Method & LiDAR & Image & Seg. & Det. & Occ. & HLF & Grounding & Zero. \\
        \hline

    \rowcolor{LGray} 3DAL~\cite{3dal} & \textcolor{ForestGreen}{\usym{2713}} & \textcolor{red}{\usym{2717}} & \textcolor{red}{\usym{2717}} & \textcolor{ForestGreen}{\usym{2713}} & \textcolor{red}{\usym{2717}} & \textcolor{red}{\usym{2717}} & \textcolor{red}{\usym{2717}} & \textcolor{red}{\usym{2717}} \\

    CTRL~\cite{fan2023once} & \textcolor{ForestGreen}{\usym{2713}} & \textcolor{red}{\usym{2717}} & \textcolor{red}{\usym{2717}} & \textcolor{ForestGreen}{\usym{2713}} & \textcolor{red}{\usym{2717}} & \textcolor{red}{\usym{2717}} & \textcolor{red}{\usym{2717}} & \textcolor{red}{\usym{2717}} \\

    \rowcolor{LGray} DetZero~\cite{ma2023detzero} & \textcolor{ForestGreen}{\usym{2713}} & \textcolor{red}{\usym{2717}} & \textcolor{red}{\usym{2717}} & \textcolor{ForestGreen}{\usym{2713}} & \textcolor{red}{\usym{2717}} & \textcolor{red}{\usym{2717}} & \textcolor{red}{\usym{2717}} & \textcolor{red}{\usym{2717}} \\

    LabelFormer~\cite{labelformer} & \textcolor{ForestGreen}{\usym{2713}} & \textcolor{red}{\usym{2717}} & \textcolor{red}{\usym{2717}} & \textcolor{ForestGreen}{\usym{2713}} & \textcolor{red}{\usym{2717}} & \textcolor{red}{\usym{2717}} & \textcolor{red}{\usym{2717}} & \textcolor{red}{\usym{2717}} \\

    \rowcolor{LGray} UniSeg~\cite{uniseg} & \textcolor{ForestGreen}{\usym{2713}} & \textcolor{ForestGreen}{\usym{2713}} & \textcolor{ForestGreen}{\usym{2713}} & \textcolor{red}{\usym{2717}} & \textcolor{red}{\usym{2717}} & \textcolor{red}{\usym{2717}} & \textcolor{red}{\usym{2717}} & \textcolor{red}{\usym{2717}} \\

    LidarMultiNet~\cite{lidar-multinet} & \textcolor{ForestGreen}{\usym{2713}} & \textcolor{red}{\usym{2717}} & \textcolor{ForestGreen}{\usym{2713}} & \textcolor{ForestGreen}{\usym{2713}} & \textcolor{red}{\usym{2717}} & \textcolor{red}{\usym{2717}} & \textcolor{red}{\usym{2717}} & \textcolor{red}{\usym{2717}} \\

    \rowcolor{LGray} SAM3D~\cite{sam3d} & \textcolor{ForestGreen}{\usym{2713}} & \textcolor{red}{\usym{2717}} & \textcolor{red}{\usym{2717}} & \textcolor{ForestGreen}{\usym{2713}} & \textcolor{red}{\usym{2717}} & \textcolor{ForestGreen}{\usym{2713}} & \textcolor{red}{\usym{2717}} & \textcolor{ForestGreen}{\usym{2713}} \\
    
    \arrayrulecolor{black}
    \cdashline{1-8}[1.5pt/4pt]

    \rowcolor{violet!10}ZOPP (ours) & \textcolor{ForestGreen}{\usym{2713}} & \textcolor{ForestGreen}{\usym{2713}} & \textcolor{ForestGreen}{\usym{2713}} & \textcolor{ForestGreen}{\usym{2713}} & \textcolor{ForestGreen}{\usym{2713}} & \textcolor{ForestGreen}{\usym{2713}} & \textcolor{ForestGreen}{\usym{2713}} & \textcolor{ForestGreen}{\usym{2713}} \\
        
        \Xhline{0.75pt}
    \end{tabular}
\end{table*}

Recently, offboard detection and auto-labeling have gained significant attention in the field of AD, which focuses on alleviating the burdens of human labor and the cost of labeling huge amounts of data.
These methods~\cite{3dal,fan2023once,ma2023detzero} have showcased impressive performance for point clouds based 3D object detection with closed-set taxonomy (\eg, predefined categories of vehicles, pedestrians, and cyclists) compared to humans.
However, their modular fashion always needs high-quality human labels as a prerequisite for training the whole pipeline, which places the auto labeling as a chicken-or-egg problem.
Due to the data sparsity and imbalance, the supervised training fashion on limited seen categories also struggles to effectively perform auto-labeling in open-set settings.
In particular, the compensated points of small and distant objects (\eg, traffic cone, traffic light) over the entire sequence are still extremely sparse, so the auto-labeling models will lose effectiveness during object-centric prediction.
Furthermore, these auto-labeling models might not flexibly generalize well due to unavoidable domain shifts arising from different types of 3D sensors.
To sum up, we found that all these shortages greatly limit the broad application prospects, and the development of a unified framework for offboard auto labeling that effectively meets the distinct needs of each perception task has not been fully explored.

To tackle this challenge, we propose ZOPP, which is a novel pioneering \textbf{Z}ero-shot \textbf{O}ffboard \textbf{P}anoptic \textbf{P}erception framework with multi-modal data input and supports a wide range of perception tasks in AD scenes.
The core of ZOPP is a compact and lightweight pipeline to achieve panoptic perception without any human-label-based model training.

Specifically, ZOPP first extends SAM-Track~\cite{cheng2023segment} to multi-view images to achieve open-set 2D detection for object tracking and instance segmentation.
Based on the aligned correspondence between point clouds and multi-view images, we can obtain robust semantic and instance segmentation for each 3D point with the proposed parallax occlusion and noise filtering module.
The points belonging to a specific object can be aggregated via the pose matrix, and then fed into the proposed point completion module to generate dense point clouds.
Equipped with such dense and high-quality object points (especially for dynamic objects), we can acquire precise 3D bounding boxes in a human-label-free manner.
Furthermore, to achieve 3D occupancy prediction, unlike straightforward voxel feature generation from image features or solely using BEV feature as in previous literature~\cite{li2023voxformer,miao2023occdepth,huang2023tri}, ZOPP employs neural rendering based reconstruction~\cite{guo2023streetsurf}  to decode 3D occupancy from the reconstructed scenes. All the instance and semantic information are fused and leads to 4D occupancy flow as the final output.

We conduct comprehensive empirical studies and evaluations on the large-scale Waymo open dataset, to validate the proposed ZOPP on various perception tasks, \ie, 2D/3D semantic and panoptic segmentation, 2D/3D detection and tracking, 4D occupancy flow prediction.
It is noteworthy that ZOPP not only produces 3D bounding boxes for the common object categories, but also integrates the open-set detection capabilities into the 3D object detection task, which shows a more profound significance for small and distant objects.
Extensive ablation studies and generalization experiments show that each proposed module of ZOPP performs well with their respective functions.

To further explore the generalization of our proposed ZOPP, we also conduct experiments in downstream applications and demonstrate ZOPP's great potential.
ZOPP can be utilized as a quick cold-start paradigm for existing auto-labeling methods. The completed dense object points can not only further boost the performance of their object-centric refining fashion, but also be used for generative assets modeling in simulation.

\section{Related Work}
\paragraph{Open-set 2D\&3D Object Detection}
Open-set object detection is trained using existing bounding box annotations and aims at detecting arbitrary classes with the help of language generalization.
Current image-based 2D open-set detectors often employ CLIP~\cite{clip} to encode the text embedding as queries to decode the category-specific boxes~\cite{ov-detr}, or as knowledge distillation to learn region embeddings containing the language semantics~\cite{vild}.
Leveraging additional data to train the model in grounding~\cite{clip-adapter} and captioning~\cite{detclip} fashions, can also improve the generalization ability.

For 3D point clouds, transferring image or vision-language pre-trained models is very challenging.
PointCLIP~\cite{pointclip} achieves open-vocabulary point-cloud recognition via projecting point-cloud into multi-view images.
Explorations of data augmentation~\cite{pointcam} and construction~\cite{3dos} are conducted to improve open-set point cloud learning.
Multi-model pre-trained models are also employed to enable open-set 3D detection for indoor scenes~\cite{ov3det,fm-ov3d}.
MLUC~\cite{mluc} combines metric learning and unsupervised clustering for limited unknown categories in the outdoors.
However, these approaches are still far from large-scale open-set settings for outdoor driving scenarios.

\paragraph{3D Segmentation and Occupancy Prediction}
3D segmentation includes semantic~\cite{cylind3d,uniseg} and panoptic~\cite{ds-net,panotpic-phnet} segmentation tasks by involving point clouds~\, or multi-modal fusion with images~\cite{zhuang2021pmf,fuseseg,madawy2019,4d-former}.
Some work~\cite{aygun2021-4dseg,zhu2023-4dseg,4d-former} also associates features from previous frames to establish 4D panoptic segmentation.
Meanwhile, zero-shot segmentation is explored by implicitly estimating the distribution of unseen features~\cite{li2023-openseg}, or visual feature guidance~\cite{lu2023smkm}.

Occupancy prediction recently arises with proposed benchmarks~\cite{tian2023occ3d,sima2023_occnet}. Visual features are leveraged to construct dense 3D occupancy with semantic labels~\cite{li2023voxformer,miao2023occdepth}.
However, these methods are all trainable with human labels or point clouds supervision~\cite{huang2023tri}.
In our offboard setting, we can employ 3D reconstruction and neural rendering~\cite{guo2023streetsurf,unisim} in our pipeline, to concentrate more on the quality of the scene geometry and visual appearance.

\paragraph{Offboard Auto Labeling}
Relying on the serialized point cloud datasets, offboard 3D detection approaches often follow a modular pipeline design~\cite{3dal,auto4d,fan2023once,ma2023detzero}, and leverage off-the-shelf 3D detectors~\cite{voxelnet,second,pointpillars,centerpoint}, trackers~\cite{ab3dmot,simpletrack,immortal_tracker}, and object-centric refining, to boost high-quality bounding boxes for auto-labeling.
LidarMultiNet~\cite{lidar-multinet} unifies 3D segmentation and detection in one network, achieving performance gains on both tasks.
Unfortunately, these methods only focus on 3D object detection, their modules still require huge amounts of data with high-quality annotations, and lack the capabilities of open-set and zero-shot settings.

In this paper, we focus on addressing the zero-shot offboard panoptic perception, and integrate the aforementioned perception tasks with an offboard running manner into the outdoor AD scenes.

\begin{figure*}[t]
\centering
    \includegraphics[width=1.\textwidth]{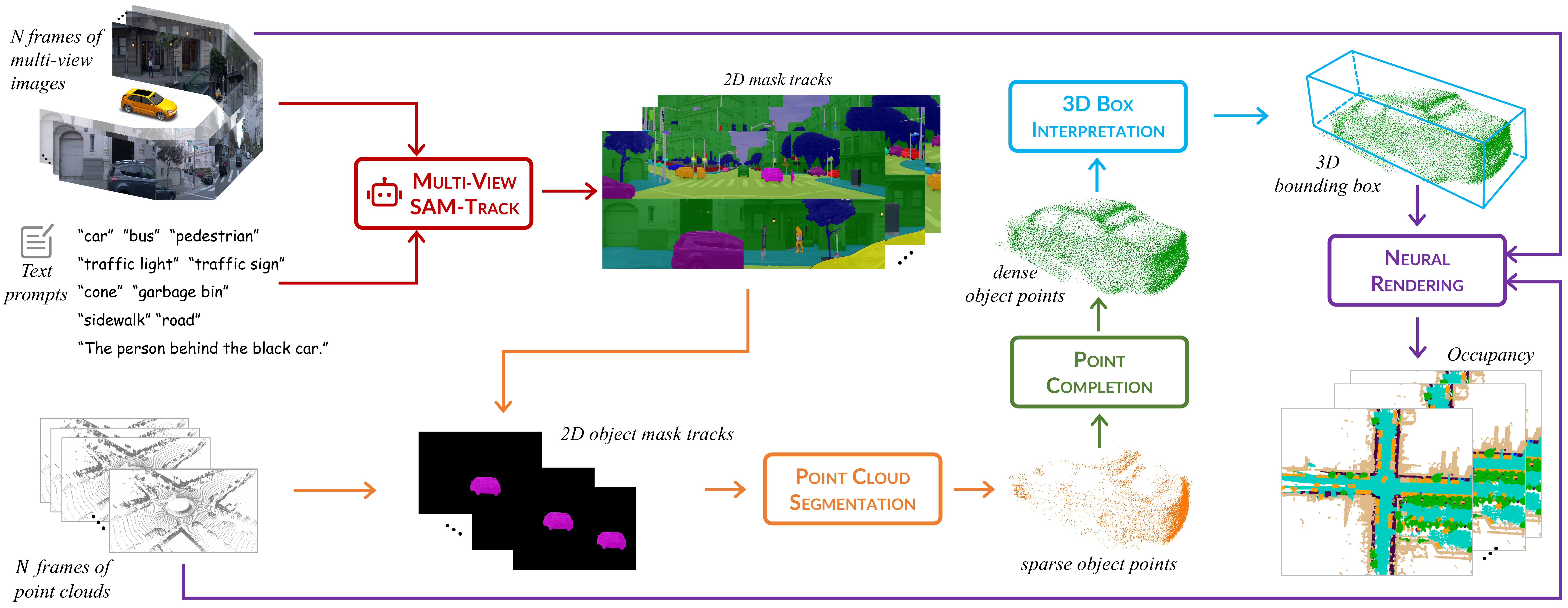}
    \caption{Overview of our proposed ZOPP.
    The core of ZOPP is a complete pipeline to achieve offboard panoptic perception of AD scenes, including multi-view mask track generation (\textbf{\textcolor{myred}{red}}), 3D semantic and instance segmentation (\textbf{\textcolor{myorange}{orange}}), point cloud completion (\textbf{\textcolor{mygreen}{green}}), 3D detection (\textbf{\textcolor{myblue}{blue}}), and 4D occupancy reconstruction (\textbf{\textcolor{mypurple}{purple}}).}
    \label{fig:framework}
\end{figure*}

\section{Methodology}
In this section, we introduce the general framework and workflow of our proposed ZOPP in detail, which generates multiple robust perception results from multi-view images and point clouds. As shown in Fig.~\ref{fig:framework}, our method comprises four stages: (1) generating multi-view object mask tracks by Multi-view SAM-Track in Sec.~\ref{sec:mv-sam-track}, (2) Point Cloud Segmentation with aligned spatial correspondence and parallax occlusion filtering in Sec.~\ref{sec:ope}, (3) 3D Box Interpretation after completing the partial points in Sec.~\ref{sec:box-itpt}, and (4) 4D Occupancy Reconstruction with neural rendering in Sec.~\ref{sec:occ-recon}.

\subsection{Multi-view Mask Track Generation}
\label{sec:mv-sam-track}
Taking as input multi-view images and text prompts, we generate 2D panoptic segmentation and tracking results with the proposed Multi-view SAM-Track.

\subsubsection{Single-view Mask Tracking}
We employ SAM-Track~\cite{cheng2023segment} to establish interactive open-set 2D object detection for segmenting and tracking in outdoor AD scenes.
Specifically, SAM-Track first includes a powerful open-set object detector, Grounding-DINO~\cite{liu2023grounding}, to detect objects in each frame according to predetermined text prompts (\eg, ``car'', ``the woman in a red dress'').
Then, SAM~\cite{sam} is leveraged to obtain segmentation masks for each object in the frame, serving as a reference input for DeAOT~\cite{yang2022deaot}, a highly efficient multi-object tracking model.
DeAOT hierarchically propagates the extracted visual embeddings and ID embeddings for each object from past to current frames based on the segmentation reference, to establish object tracking frame-by-frame.

\subsubsection{Multi-view SAM-Track}
Considering the prevalent use of multi-view cameras in AD, we design a simple yet effective similarity cost to measure the semantic and instance consistency among objects across all the views. This cost involves the computation of appearance and location similarities to facilitate object association.

We first apply the aforementioned process to each image sequence of different views, yielding independent tracking results.
Simultaneously, we obtain the appearance information of each object by extracting the visual features of Grounding-DINO and DeAOT with the 2D boxes.
So the appearance similarity is compared across different objects by computing the cosine distance of the visual features.
In contrast, the location similarity is derived by concatenating the images of all viewpoints in a panoramic order, followed by normalizing the pixel distances along the horizontal axis for each object.
Hence, objects with large similarity scores would be associated together with the same instance ID.

\begin{figure*}[]
\centering
\includegraphics[width=1.\textwidth]{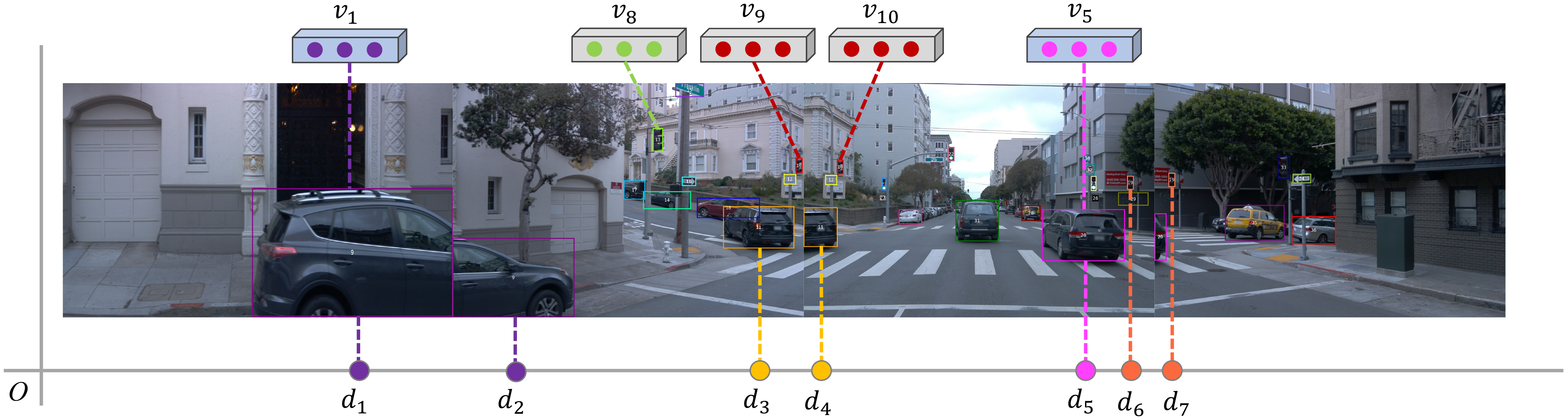}
\caption{Overview of our object association across multiple views. Multi-view images are concatenated in a panoramic order. The visual features and horizontal pixel coordinates of each object are drawn at the top and bottom of the images, respectively. Visual features $v_1$ and $v_5$ are very similar, so the location differences $d_1$ and $d_5$ contribute to the matching determination. The visual features of traffic lights are almost the same ($v_8, v_9, v_{10}$), so we can associate them with location similarities ($d_6,d_7$).}
\label{fig:mv-sam-track}
\end{figure*}

As illustrated in Fig.~\ref{fig:mv-sam-track}, the use of appearance similarity allows for the discrimination of objects that are spatially close but exhibit significant visual differences.
Meanwhile, distance similarity serves to prevent the matching of objects with similar appearances but substantial spatial separation.
This comprehensive design thereby enhances the robustness and accuracy in multi-view settings.

For the sake of the grounding ability, we preserve the interactive mode to select target objects through natural language. For the automatic mode, we output all the object categories arising at the driving surroundings, to establish multi-view panoptic segmentation.
Finally, we directly output the tracked object masks with a unique ID and corresponding categories as the final 2D semantic and instance segmentation results.
Note that background objects (\eg, buildings, trees) only have semantic segmentation results.

\subsection{Point Cloud Segmentation}
\label{sec:ope}
In this section, point cloud data is introduced to be well-aligned with multi-view image planes to obtain corresponding semantic and instance information.
Then, we can extract points belonging to each foreground object based on the instance ID, for subsequent 3D box interpretation.
The extraction is carefully established by our proposed parallax occlusion and noise filtering module.

\subsubsection{Multi-modal Spatial Alignment}
We denote a frame of point cloud as $\mathcal{P}^\mathrm{L} = \{p_1^\mathrm{L}, p_2^\mathrm{L}, ...\}$, where $\mathrm{L}$ represents the LiDAR coordinate system.
For each 3D point $p_i^\mathrm{L}=(x_i,y_i,z_i)^T \in \mathbb{R}^3$, we denote its corresponding pixel coordinate on the image plane as $q_i = (u_i,v_i)^T \in \mathbb{R}^2$. 
The point and the pixel can be correlated by the calibration process in two steps.
Firstly, $p_i^\mathrm{L}$ is transformed to the camera coordinate system $\mathrm{C}$ as  $p_i^\mathrm{C} \in \mathbb{R}^3$ through $p_i^\mathrm{C} = \boldsymbol{\mathrm{R}} \cdot p_i^\mathrm{L} + \boldsymbol{\mathrm{t}}$ ($\boldsymbol{\mathrm{R}}$ and $\boldsymbol{\mathrm{t}}$ represent the rotation and translation between LiDAR and multi-view cameras).
Next, $p_i^\mathrm{C}$ is projected onto the image plane through a projection function: $q_i = \boldsymbol{\mathrm{K}} (p_i^\mathrm{C})$ ($\boldsymbol{\mathrm{K}}: \mathbb{R}^3 \rightarrow \mathbb{R}^2$ is defined with the camera intrinsic parameter for each specific view).

\subsubsection{Parallax Occlusion and Noise Filtering}
\begin{figure*}[htbp]
\centering
\includegraphics[width=1.\textwidth]{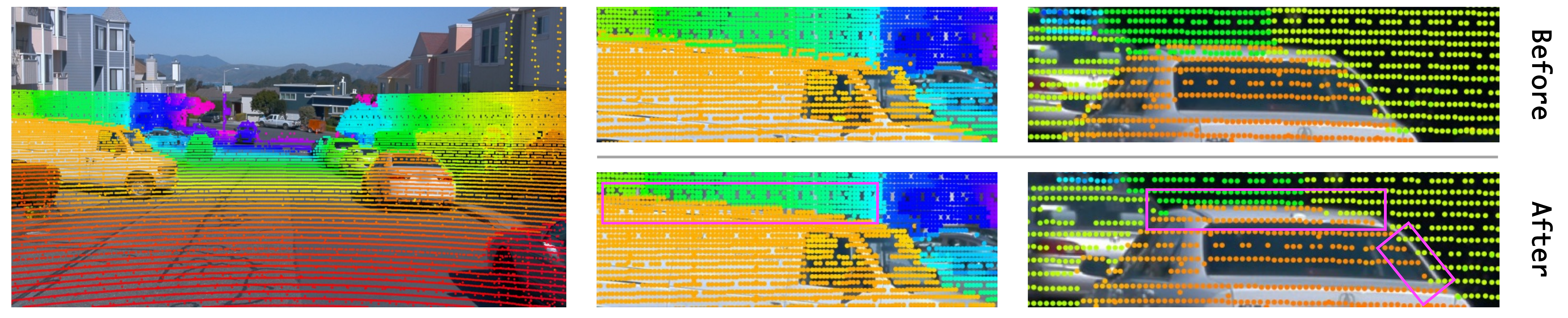}
\caption{
Point clouds are projected into the image plane, and visualized in a color map based on the depth values (\textbf{\textcolor{red}{Near}} to \textbf{\textcolor{blue}{Far}).}
On the right, we compare the effect before (top) and after (bottom) our proposed parallax occlusion. Please zoom in the highlighted pink boxes to see the filtering points.}
\label{fig:vis-abl-parallax}
\end{figure*}

Given well-aligned point-to-pixel correspondence, we can easily obtain the instance ID and semantic categories for most projected 3D points within the 2D object masks.
This strategy is leveraged by most of the previous methods to obtain 3D mask annotations~\cite{openannotate3d}.
However, LiDARs are always equipped much higher than multi-view cameras on autonomous vehicles, leading to serious parallax occlusion issues.
As shown in Fig.~\ref{fig:vis-abl-parallax}, the 3D points belonging to backgrounds (green) are projected into the pixel regions of the car (orange).
Because disparity occlusion commonly arises at regions around the upper edges of foreground objects, we thereby propose to filter out these background points from the foreground pixels with an algorithm akin to a convolution filtering operation.

\begin{figure*}[htbp]
    \begin{minipage}{0.58\textwidth}
        \begin{algorithm}[H]
        \small
        \caption{Parallax Occlusion Filtering}
        \label{alg:parallax}
        \SetAlgoLined
        \KwIn{projected points $\mathcal{P}^{\mathrm{I}}$, 2D instance segmentation masks $\mathcal{M}$, kernel size $k$, horizontal and vertical step size $s_h, s_v$, depth threshold $\theta$, image resolution $h, w$}
        \KwOut{accurate object-specific points $\mathcal{P}^{\mathrm{L}}_i$} 
        \BlankLine
        \For{$\mathcal{M}_i \leftarrow$ near to far}{
            $\mathrm{cnt\_h}=0, \mathrm{cnt\_w}=0$\;
            $\mathcal{P}^\mathrm{I} \leftarrow$ \texttt{SpatialAligment($\mathcal{M}_i, \mathcal{P}^\mathrm{L}$)}\;
            \While{$\mathrm{cnt\_h} < h, \mathrm{cnt\_w} < w$}{
                $p \leftarrow$ \texttt{SampleDepthPixel($\mathcal{P}^\mathrm{I}$, $\mathrm{cnt\_h}$, $\mathrm{cnt\_w}$, $k$)}\;
                \If{$\frac{\max(p)-\min(p)}{\min(p)}>\theta$}{
                    $p^\text{near}, p^\text{far} \leftarrow$ \texttt{SplitDepthPixel($p$, $\theta$)}\;
                    \lIf{$\mathrm{len(p^\text{near})}>1$}{$\mathrm{Rect} \leftarrow$ \texttt{LocalRectConstruct()}}
                    \lElse{$\mathrm{Rect} \leftarrow$ \texttt{LocalRectConstruct()}}
                    \If{$p^\text{far}$ in $\mathrm{Rect}$}{$\mathcal{P}^{\mathrm{L}}_i \leftarrow$\texttt{FilterOut($p^\text{far}$)}}
                }
                $\mathrm{cnt\_h}$+=$s_h$, $\mathrm{cnt\_w}$+=$s_w$
            }
        }
    \end{algorithm}
    \end{minipage}
    \hspace{3mm}
    \begin{minipage}{0.35\textwidth}
        \centering
        \includegraphics[width=0.8\textwidth]{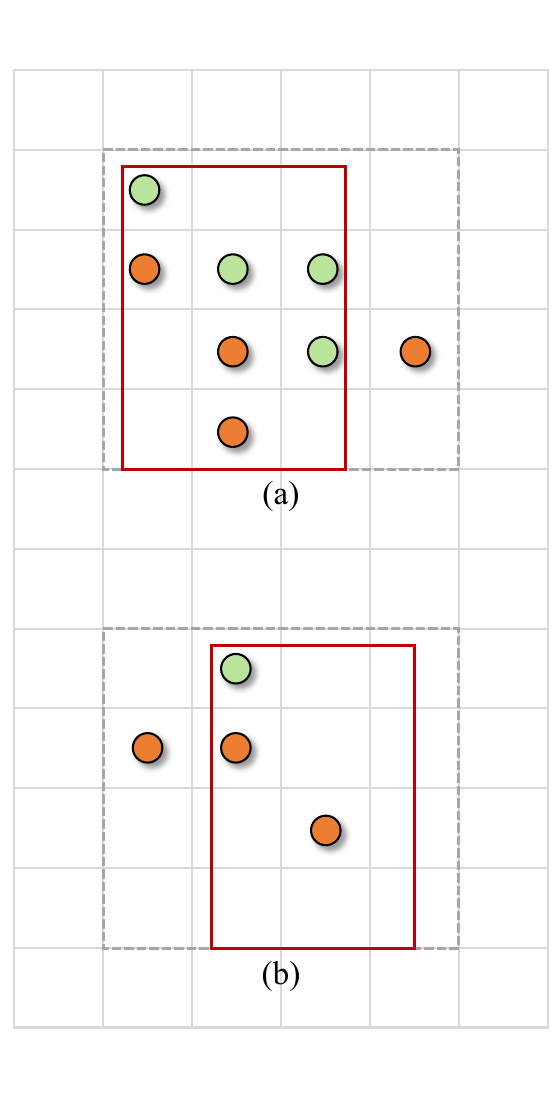}
        \vspace{-3mm}
        \caption{Two cases of constructing the local rectangle regions in our proposed algorithm. Projected points with large depth values $p^\text{far}$ (orange) inside the local rectangle region will be filtered out.}
        \label{fig:parallax-kernel}
    \end{minipage}
\end{figure*}

In specific, we first take a subset of the projected points that fall within a pre-defined filter kernel, and then calculate the maximum depth differences among them.
If the depth difference exceeds a threshold $\theta$, we identify all points that surpass $\theta$ as $p^\text{far}$, and $p^\text{near}$ otherwise.
Then, we will construct a local rectangle region $\boldsymbol{\mathcal{R}}$ based on the numbers of $p^\text{near}$.

As illustrated in Fig.~\ref{fig:parallax-kernel}, there are two cases to be considered: (1) If the number of $p^\text{near}$ is greater than one, we identify the maximum and minimum pixel coordinates along the horizontal axis among $p^\text{near}$ to define the left and right boundaries of $\boldsymbol{\mathcal{R}}$, and (2) if there is only one $p^\text{near}$, we record its coordinate as the left boundary and assume a pseudo coordinate along the horizontal direction as the right boundary.
Meanwhile, the top and bottom boundaries are determined by the minimum vertical pixel coordinate among $p^\text{near}$ and the maximum vertical coordinate of the location covered by the operation kernel, respectively.
Finally, $p^\text{far}$ that fall within $\boldsymbol{\mathcal{R}}$ will be filtered out as background points to be assigned with relevant semantic categories (\eg, road, sidewalk, wall).

In practice, we execute the filtering operation for each object mask after sorting their depth values in ascending order. The kernel size and stride of this operation kernel can be adjusted to different LiDAR types. Detailed steps are presented in Alg.~\ref{alg:parallax}.
Note that the projected points may be located in two valid 2D object masks from neighboring views. Thanks to our multi-view consistency design, we could directly combine the object points together with the same instance ID.
In addition, we filter out isolated outliers and noise points by clustering the re-projected object points in 3D space with DBSCAN~\cite{dbscan} technique.

With the processing of all these methods, we can extract points belonging to foreground objects and assign accurate semantic categories and instance IDs for them, resulting in the final 3D semantic and instance segmentation outputs.

\subsection{3D Box Interpretation}
\label{sec:box-itpt}
In this section, we aim to interpret the precise 3D bounding boxes with the instance segmented points in a human-label-free manner, especially for the foreground objects.

However, super sparse point clouds of objects are very common in driving environments, typically manifesting in two scenarios: (1) LiDARs often struggle to obtain dense scanning results for small or distant objects, and (2) due to the constraints of fixed configurations, it becomes challenging to capture scans of objects from all views.
Hence, it is challenging to precisely characterize the geometry shape even after compensating the object points of the entire sequence.
To meet the requirement of deprecating human labels, it is intuitive to first complete point clouds from partial inputs.

\subsubsection{Point Completion}
Inspired by recent remarkable progress in the field of point cloud completion, we design a simple and effective network to precisely capture the structural information of 3D shapes and predict complete point clouds with highly detailed geometries.

The whole network consists of three models, the point encoder, geometry generator, and point generator.
Specifically, a PointNet-structure~\cite{pointnet} encoder first extracts a shape embedding from the partial point cloud input to capture both local structural details and the global context of the object.
To better take advantage of semantic information, we leverage pre-trained CLIP~\cite{clip} text encoder to generate object category embedding.
Then, the geometry generator aims to produce a sparse but complete geometric structure, based on decoding the shape and category embeddings.
The final point generator receives the shape embedding, the geometric structure, and the category embedding as input, and generates the dense fine-grained point clouds.
All the point cloud modules leverage self-attention layers to adaptively aggregate information and reveal detailed spatial relations among the unordered partial points.
Please refer to the Appendix for more details on acquiring partial-complete data pairs and the model training process.

\subsubsection{Box Interpretation}
We first classify the motion state of the objects as static or dynamic based on the segmented points of each object track.
For static objects, we transform the object points of each frame to the global coordinate with the pose matrix, and combine them together. We then apply L-Shape fitting to derive an initial 3D box representing the geometric shape. Noisy points and outliers outside the initial box are removed, and we randomly select a set of points with FPS sampling.
L-Shape fitting is then performed for these selected points to generate a refined 3D box, which is subsequently transformed back to each frame as the final result.

In contrast to the combination operation for static objects, we process each sample of the dynamic object tracks on a frame-by-frame basis. We first subsample a set of points with FPS sampling and fit the initial 3D box for each object sample from each frame.
Based on the distributions of these initial boxes, we generate anchors that stabilize the refined 3D box through L-Shape fitting.
Finally, the trajectory is smoothed by linear fitting and Kalman filter in the global coordinate, and the results are then updated to each frame.

\subsection{4D Occupancy Flow}
\label{sec:occ-recon}
Eventually, multi-view images, point clouds, and the generated 3D boxes are all fed into a neural rendering model to reconstruct the 3D scenes, which are used to decode occupancy grids as our 4D occupancy flow output.

In particular, we aim to build a compositional scene representation that models the 3D world including the dynamic objects and static scene, by leveraging StreetSurf~\cite{guo2023streetsurf}.
The core is to render the well-suit geometry representation with signed distance functions (SDF), by disentangling a 3D space volume into a static background and a set of foreground objects~\cite{unisim,drivinggaussian} which are determined by the input 3D boxes.
Please refer to the Appendix or the original paper for more details.

With the implicit surface being reconstructed, we obtain a continuous representation of scene geometry that has infinitesimal granularity. Subsequently, we can decode high-resolution occupancy grids out of the reconstructed implicit surface.
The semantic and instance information of each grid can still be preserved based on the inside LiDAR points.

\section{Experiments}
In this section, we first introduce the dataset details and evaluation metrics. We then provide a detailed performance of ZOPP on different perception tasks.
The ablation studies and analysis are presented to convince each component of our entire approach.
Please refer to Appendix for detailed quantitative results, more qualitative results, and application experiments.

\subsection{Dataset}
Following the experimental setting of previous offboard perception methods~\cite{3dal,ma2023detzero,fan2023once,labelformer}, we conduct extensive experiments on the large-scale Waymo open dataset~\cite{sun2020scalability}. The dataset provides 20-second point cloud and 5-view image data for each scene with a sampling frequency at 10Hz.
Considering that the environmental conditions would affect the quality of neural rendering (\eg, weather conditions, image blurring), we select a set of sequences from the validation set to conduct all the experiments.

\subsection{Main Results}
We present a comprehensive evaluation of 3D object detection, 3D segmentation, and occupancy prediction.
Note that there are only 5 cameras on WOD, we hence calculate the performance (indicated by $\dagger$) of each perception task by excluding regions outside the field of view of multiple cameras.

\begin{figure*}[ht]
\centering
\includegraphics[width=0.95\textwidth]{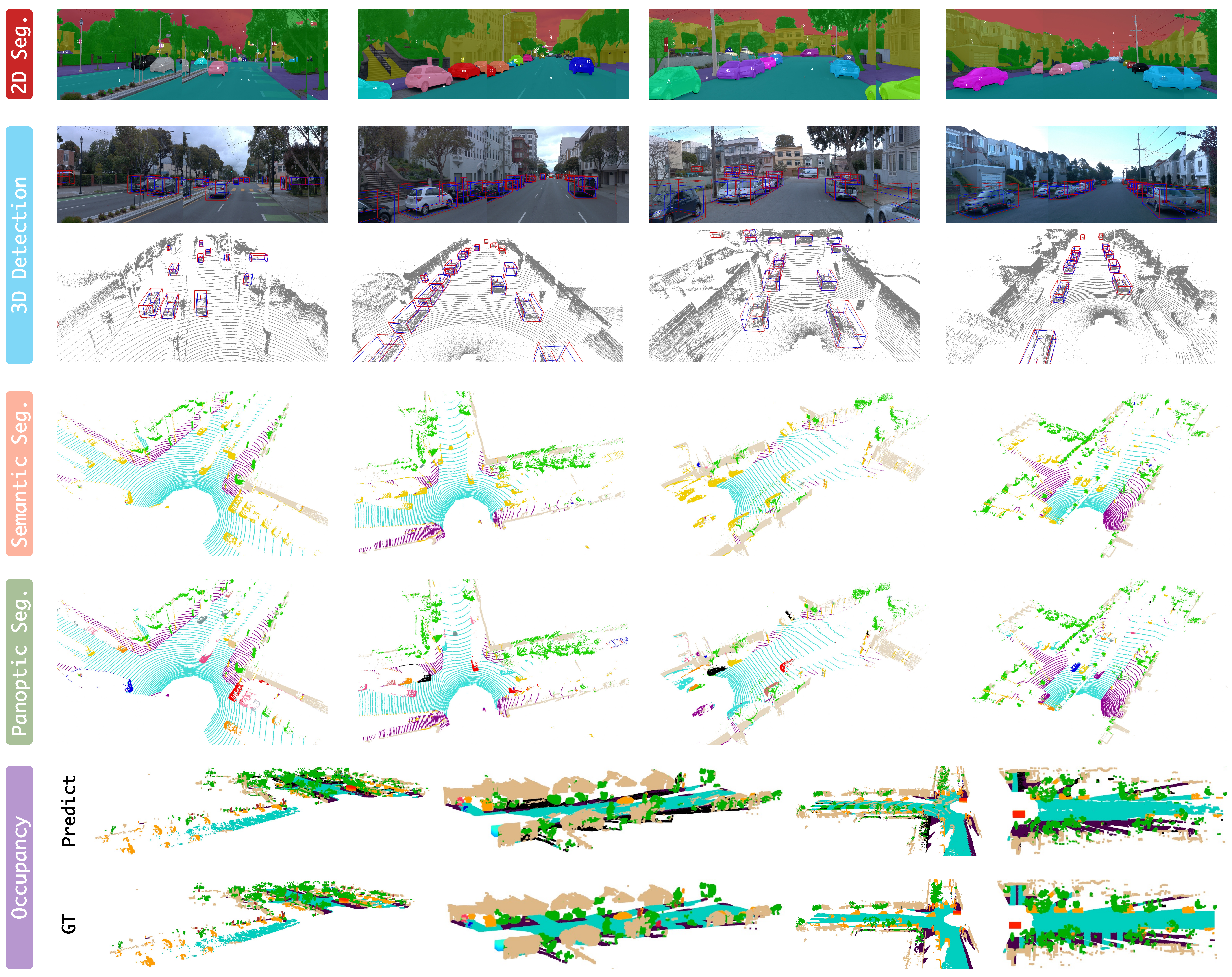}
\caption{Qualitative results of our proposed ZOPP on various perception tasks in AD scenes, including \textbf{\textcolor{myred}{2D segmentation}}, \textbf{\textcolor{myblue}{3D detection}}, \textbf{\textcolor{myorange}{3D semantic segmenation}},
\textbf{\textcolor{mygreen}{3D panoptic segmentation}},
and \textbf{\textcolor{mypurple}{occupancy predition}}.}
\label{fig:vis-all}
\end{figure*}

\paragraph{3D Detection}
As illustrated in Tab.~\ref{tab:res-det}, we report the performance of our ZOPP on the validation set of WOD.
The Average Precision (AP) and Recall performance are calculated using different matching criteria (IoU and BEV distance).
Meanwhile, we compare the performance with several methods across different distance ranges in Tab.~\ref{tab:res-det-compare}.
As the distance increases, the performance of all methods decreases. Specifically, VoxelRCNN shows a decline in L1 AP of 14.37\% and 36.12\% for the distance ranges of \textit{30-50m} and \textit{50+m}, compared to \textit{0-30m}. PVRCNN experiences decreases of 15.52\% and 37.05\%, while our method demonstrates reductions of 16.94\% and 29.35\%.
This improvement can be attributed to our mask tracking module, which effectively utilizes the entire temporal information in the point cloud sequence with generated object IDs. Consequently, our method mitigates the impact of distance more effectively than other onboard methods, particularly at farther ranges.
Furthermore, we visualize the 3D object detection results in Fig.~\ref{fig:vis-all}, where the red and blue boxes are ground-truth and predicted ones, respectively.

\begin{table}[htbp]
\small
\setlength{\tabcolsep}{0.32cm}
\caption{Verifying 3D object detection ability of our ZOPP on WOD val set. Metrics are 3D AP of L2 difficulties for \textit{Vehicle}, \textit{Pedestrian}, and \textit{Cyclist}.}
\label{tab:res-det}
\begin{center}
\renewcommand{\arraystretch}{1.2}
  \begin{tabular}{l|cc|cc|cc}
    \Xhline{0.75pt}
     & \multicolumn{2}{c|}{\textit{Vehicle}} & \multicolumn{2}{c|}{\textit{Pedestrian}} & \multicolumn{2}{c}{\textit{Cyclist}} \\
    
    {Criterion} & {AP} & {Recall} & {AP} & {Recall} & {AP} & {Recall} \\
    
    \hline

    {IoU$^{\dagger}$} & {$35.6$} & {$48.8$} & {$34.5$} & {$46.7$} & {$11.2$} & {$22.9$} \\

    % \hdashline
    {Distance$^{\dagger}$} & {$48.1$} & {$61.6$} & {$46.7$} & {$58.5$} & {$21.8$} & {$34.0$} \\
    
    \Xhline{0.75pt}
  \end{tabular}
  \end{center}
\end{table}

\begin{table}[htbp]
\small
\setlength{\tabcolsep}{0.2cm}
\caption{Comparisons of fully-supervised detectors and human-label-free methods. We re-implement these methods and report the AP performance (IoU criterion) of \textit{Vehicle} within camera FOVs across different distance ranges.}
\label{tab:res-det-compare}
\begin{center}
\renewcommand{\arraystretch}{1.05}
  \begin{tabular}{l|c|cc|cc|cc|cc}
    \Xhline{0.75pt}
    &  & \multicolumn{2}{c|}{\textit{Total}} & \multicolumn{2}{c|}{\textit{0-30m}} & \multicolumn{2}{c|}{\textit{30-50m}} & \multicolumn{2}{c}{\textit{50+m}} \\
    
    {Method} & {Training Data} & {L1} & {L2} & {L1} & {L2} & {L1} & {L2} & {L1} & {L2} \\
    
    \hline

    {Centerpoint~\cite{centerpoint}} & {\textit{train set}} & {$73.04$} & {$64.72$} & {$88.17$} & {$86.81$} & {$72.12$} & {$66.50$} & {$51.24$} & {$39.72$} \\

    {VoxelRCNN~\cite{voxelrcnn}} & {\textit{train set}} & {$76.29$} & {$67.05$} & {$89.27$} & {$87.84$} & {$76.44$} & {$69.68$} & {$57.03$} & {$44.37$} \\

    {PVRCNN~\cite{pvrcnn}} & {\textit{train set}} & {$75.53$} & {$66.77$} & {$89.03$} & {$87.63$} & {$75.21$} & {$68.33$} & {$56.04$} & {$43.33$} \\

    {DetZero~\cite{ma2023detzero}} & {\textit{train set}} & {$89.49$} & {$83.34$} & {$96.64$} & {$95.90$} & {$88.84$} & {$84.37$} & {$78.32$} & {$66.77$} \\

    {SAM3D~\cite{sam3d}} & {-} & {$6.90$} & {$5.88$} & {$19.51$} & {$19.05$} & {$0.029$} & {$0.026$} & {$0.0$} & {$0.0$} \\

    {ZOPP$^\dagger$ (ours)} & {-} & {$37.56$} & {$35.61$} & {$42.31$} & {$41.16$} & {$35.14$} & {$33.86$} & {$29.89$} & {$28.67$} \\
    
    \Xhline{0.75pt}
  \end{tabular}
  \end{center}
\end{table}

\paragraph{Segmentation}
We present the results of semantic segmentation and panoptic segmentation for both 2D images and 3D point clouds in Fig.~\ref{fig:vis-all}. In addition to common objects such as vehicle, the segmentation results for tree, pole, traffic light, and sign, are also impressive. This demonstrates that we not only retain the dense semantic and instance information from the foundation models, but also establish carefully aligned correspondence by the proposed parallax occlusion and noise filtering.
The quantitative results of our ZOPP are shown in Tab.~\ref{tab:app-seg-miou}, along with the performance of SOTA methods for reference.
We achieve comparable performance, particularly on foreground objects (\eg, Vehicle, pedestrian, bicyclist), showcasing the powerful potentials of our point clouds segmentation module.
Note that we merge all the categories belonging to \textit{car, truck, bus, other vehicle} together as \textit{Vehicle} (its performance is the average of these four 
categories). Additionally, categories that cannot be fully recognized are excluded from the results.

\begin{table*}[htbp]
\caption{Comparisons of ZOPP and state-of-the-art LiDAR semantic segmentation methods.}
\label{tab:app-seg-miou}
\setlength{\tabcolsep}{0.004\linewidth}
\centering
\begin{adjustbox}{width=0.99\columnwidth,center}
\begin{tabular}{l| c  c c c c c c c c c c c c c c c c c c}
    \toprule
    Method 
    & \rotatebox{90}{Vehicle} 
    & \rotatebox{90}{motorcyclist}
    & \rotatebox{90}{bicyclist} 
    & \rotatebox{90}{pedestrian} 
    & \rotatebox{90}{sign} 
    & \rotatebox{90}{traffic light} 
    & \rotatebox{90}{pole} 
    & \rotatebox{90}{Cons. Cone} 
    & \rotatebox{90}{bicycle} 
    & \rotatebox{90}{motorcycle} 
    & \rotatebox{90}{building} 
    & \rotatebox{90}{vegetation} 
    & \rotatebox{90}{tree trunk}
    & \rotatebox{90}{curb}
    & \rotatebox{90}{road} 
    & \rotatebox{90}{lane marker}
    & \rotatebox{90}{other ground}
    & \rotatebox{90}{walkable}
    & \rotatebox{90}{sidewalk} \\
    \midrule

    P-Transformer~\cite{ptrans}
    & 59.7 & 0.0 & 67.9 & 85.5 & 72.3 & 36.2 & 71.4 & 66.4 & 58.7 & 54.3 & 93.7 & 90.0 & 64.7 & 65.2 & 90.4 & 48.2 & 42.8 & 74.5 & 71.7 \\

    UniSeg~\cite{uniseg} & 68.8 & 0.0 & 73.2 & 89.0 & 75.7 & 43.3 & 76.1 & 70.2 & 75.5 & 80.8 & 95.2 & 91.0 & 68.2 & 68.7 & 92.6 & 53.9 & 48.3 & 78.8 & 75.8 \\

    ZOPP$^\dagger$ (ours) & 54.2 & - & 49.6 & 77.3 & 29.7 & 34.2 & 51.7 & 33.1 & 21.8 & 35.4 & 75.5 & 73.6 & - & - & 81.8 & - & - & - & 61.2 \\

\bottomrule
\end{tabular}
\end{adjustbox}
\end{table*}

\paragraph{Occupancy}
We leverage Occ3D~\cite{tian2023occ3d} as ground truth to evaluate the performance. Our approach is based on neural rendering reconstruction, which operates in an offboard fashion, prioritizing reconstruction quality over running efficiency. The performance of other training-based methods is reported as a reference, rather than for detailed comparison.
As shown in Tab.~\ref{tab:occ-res}, we achieve promising results compared to previous methods, especially for slim objects (\eg, traffic light, pole) and flat objects (\eg, road). Object categories not recognized in the selected sequences are excluded (\eg, tree trunk).
Notably, ZOPP can outperform a LiDAR-only baseline (supervised training) in some object categories with limited training samples (\eg, traffic light, sign, cone), by utilizing the zero-shot capabilities of foundation models.
Meanwhile, as shown in Fig.~\ref{fig:vis-all}, the reconstructed scenes are highly complete and spatially coherent, \eg, the predicted road is highly complete and accurately well-defined, demonstrating that our reconstruction method can preserve the detailed 3D geometry effectively.

\begin{table*}[!h]
\caption{Comparison of 3D occupancy prediction performance.}
\label{tab:occ-res}
\setlength{\tabcolsep}{0.004\linewidth}
\centering
\begin{adjustbox}{width=0.99\columnwidth,center}
\begin{tabular}{l| c  c c c c c c c c c c c c c c|c}
    \toprule
    Method 
    & \rotatebox{90}{GO} 
    & \rotatebox{90}{vehicle}
    & \rotatebox{90}{bicyclist} 
    & \rotatebox{90}{pedestrian} 
    & \rotatebox{90}{sign} 
    & \rotatebox{90}{traffic light} 
    & \rotatebox{90}{pole} 
    & \rotatebox{90}{Cons. Cone} 
    & \rotatebox{90}{bicycle} 
    & \rotatebox{90}{motorcycle} 
    & \rotatebox{90}{building} 
    & \rotatebox{90}{vegetation} 
    & \rotatebox{90}{tree trunk} 
    & \rotatebox{90}{road} 
    & \rotatebox{90}{sidewalk} 
    & \rotatebox{90}{mIoU}  \\
    \midrule

    TPVFormer~\cite{huang2023tri}   & 3.89 & 17.86 & 12.03 & 5.67 & 13.64 & 8.49 & 8.90 & 9.95 & 14.79 & 0.32 & 13.82 & 11.44 & 5.8 & 73.3 & 51.49 & 16.76\\

    BEVFormer~\cite{li2022bevformer}  & 3.48 & 17.18 & 13.87 & 5.9 & 13.84 & 2.7 & 9.82 & 12.2 & 13.99 & 0.0 & 13.38 & 11.66 & 6.73 & 74.97 & 51.61 & 16.76 \\

    BEVFormer-Fusion & 5.11 & 64.61 & 52.35 & 21.52 & 32.74 & 17.1 & 42.62 & 27.75 & 13.36 & 0.05 & 63.65 & 60.51 & 35.64 & 81.89 & 66.84 & 39.05 \\

    LiDAR-only & 1.01 & 57.41 & 35.31 & 20.33 & 11.7 & 13.01 & 36.21 & 7.81 & 0.13 & 0.0 & 57.83 & 54.71 & 27.07 & 69.15 & 54.47 & 29.74 \\

    ZOPP$^\dagger$ (ours) & 0.08 & 49.68 & 10.63 & 6.44 & 12.33 & 21.73 & 32.75 & 19.87 & 9.41 & 0.07 & 41.14 & 46.22 & - & 69.07 & 32.34 & 25.13 \\

\bottomrule
\end{tabular}
\end{adjustbox}
\end{table*}

\section{Conclusion}
In this work, we have proposed ZOPP, a novel framework of zero-shot offboard panoptic perception for autonomous driving.
Foundation models empower our ZOPP comprehensive capability of language understanding to establish various perception tasks for open-set settings in a zero-shot manner.
We enhance SAM-Track to ensure semantic and instance consistency among object mask tracks across multiple views.
The proposed parallax occlusion and noise filtering can produce robust 3D semantic and panoptic segmentation results after the well-aligned correspondence between point clouds and multi-view image planes.
Equipped with the proposed point completion module, we can generate dense completed points and subsequently interpret precise 3D bounding boxes.
These modules cooperate to make the 3D segmentation and detection more accurate and consistent, especially for dynamic foreground objects.
Finally, we decode high-quality 4D occupancy by concentrating on the geometry quality and visual appearance with neural rendering reconstruction fashion.
Extensive experimental results not only demonstrate that ZOPP substantially advances promising open-set perception results in offboard manner for outdoor AD scenes, but also show a profound significance in industry auto-labeling applications.

\section{Limitations and Broader Impacts}
\label{sec:limit}
While foundation models have endowed our ZOPP with open-set capabilities, the annotated categories in the existing dataset still incorporate expressions that lack universality, which may hinder the effective recognition of similar object categories.
Additionally, neural rendering may encounter numerous challenges in street-view scenes, influenced by practice factors (adverse weather conditions, sensor imaging issues).
Moreover, ZOPP may raise concerns about data capturing, abuse, privacy, and legal implications in driving surroundings.
Nonetheless, ZOPP still offers a high degree of flexibility, allowing for seamless integration with SOTA models to meet diverse application requirements, showcasing resilience and applicability in both industry and daily lives.
We believe that advancements in technology and the development of regulatory frameworks can pave the way for unified AD systems.

\section*{Acknowledgements}
This project is funded in part by National Key R$\&$D Program of China Project 2022ZD0161100, by Shanghai Artificial Intelligence Laboratory (Grant No. 2022ZD0160104), by the Centre for Perceptual and Interactive Intelligence (CPII) Ltd under the Innovation and Technology Commission (ITC)’s InnoHK, by General Research Fund of Hong Kong RGC Project 14204021, by Smart Traffic Fund PSRI/76/2311/PR for algorithm framework design and dataset curation. Hongsheng Li is a PI of CPII under the InnoHK.

\small{
\bibliographystyle{unsrt}
\bibliography{egbib}
}

%%%%%%%%%%%%%%%%%%%%%%%%%%%%%%%%%%%%%%%%%%%%%%%%%%%%%%%%%%%%
\newpage
\appendix
\normalsize
\section*{Appendix}

\section{Overview}
This document is the supplementary material of our ZOPP. We provide more details of models, experiments and analysis results in this document.

Sec.~\ref{sec:app-pcc-data} provides the details of acquiring partial-complete data pairs and model training process.
Sec.~\ref{sec:app-nerf} introduces more details about the neural rendering reconstruction method.
Sec.~\ref{sec:app-implent} shows the implementation details of the network setting and training process.
Sec.~\ref{sec:app-exp} provides more experiment results and analyses in detail. In specific, we compare the distribution of predicted bounding boxes with different distance thresholds in Sec.~\ref{sec:app-exp-det}.
The effectiveness of our method to overcome the influence of occlusion is shown in Sec.~\ref{sec:app-occlusion}.
The qualitative and quantitative results to show the effectiveness of parallax occlusion and noise filtering are presented in Sec.~\ref{sec:app-parallax}.
And the effectiveness of point completion in our whole pipeline is shown in Sec.~\ref{sec:app-exp-pcc}.
We also illustrate the open-set detection capabilities in Sec.~\ref{sec:app-openset}, and the failure pattern analysis in Sec.~\ref{sec:app-failure}.

\section{Method Details}
\subsection{Point Completion}
\label{sec:app-pcc-data}
We implement an automatic algorithm to determine whether the extracted object points are completed or not, selectively filtering out those with intact shapes.
The automatic selection is mainly based on the ratio of occupied grids.
During the training process, we randomly remove part of the points to generate the partial inputs, based on the constraints of geometric projection principles, inducing realistic structural incompleteness data pairs.
For the sparse but complete geometric points, we randomly sample a set of points based on the FPS sampling strategy to promise the geometric structure.
We also use random rotations sampled from a uniform distribution $[-\pi/2, \pi/2]$ and random linear transformations sampled from a standard Gaussian distribution to translate the point coordinates.
Chamfer distance is utilized as the metric distance of points to formulate the supervision between partial input and real dense complete object point clouds.
We don't require any human labels in this procedure.

\subsection{Neural Rendering}
\label{sec:app-nerf}
We aim to build a compositional scene representation that models the 3D world including the dynamic objects and static scene, by leveraging a neural rendering method~\cite{guo2023streetsurf}.
A 3D space volume is first defined over the entire scene. The volume consists of a static background and a set of dynamic objects, determined by the input 3D boxes. Such that separate neural feature fields and feature grids can be used to model them, respectively.

The static background is delimited into three parts, close-range, distant view, and sky. This design can well-suit geometry representation by signed distance functions (SDF). Then, three neural rendering models are employed for these three parts to jointly render a differential pixel color by querying samples for each ray. The queried samples are combined from near to far for the subsequent volume rendering.

The dynamic foreground objects are transformed to their local box coordinates (centroid of the box), and their feature grids are at the world coordinate to compose with the background. allowing us to disentangle the 3D motion of each object and focus on representing shape and appearance~\cite{unisim,drivinggaussian}.

\section{Implementation Details}
\label{sec:app-implent}
For parallax occlusion filtering, the kernel size is $15$, the steps in horizontal and vertical directions are $10$ and $5$ respectively. The depth ratio threshold is set to $0.25$.
For Grounding-DINO, we keep the same setting to leverage pre-trained Swin-L~\cite{swin} as image backbone, and BERT-base~\cite{bert} from Hugging Face~\cite{hugface} as text backbones.
The point completion network is tuned on WOD training set, with our proposed data preparation mentioned before. The whole training includes 100 epochs because of the limited amounts of objects, while the learning rate is initialized to $1e-4$ and decayed by $0.7$ every 40 epochs with Adam optimizer. The batch size is set to 32.
For occupancy reconstruction, we train the model for around $15000$ iterations and add additional cross-entropy supervision after $5000$ iterations, compared to the original version~\cite{guo2023streetsurf}. In one batch, we use $8192$ rays. The entire pipeline does not rely on too many computation resources, the point completion module and the reconstruction module need to train the network. We utilize four NVIDIA A100 to accelerate the reconstruction with multi-processing settings.

\section{More Experiments and Ablations}
\label{sec:app-exp}

\subsection{3D Detection Analysis}
\label{sec:app-exp-det}
In this section, we show a detailed analysis of our 3D detection performance shown in Tab.~\ref{tab:app-det-recall}.
Specifically, we first match the predicted boxes with the ground-truth boxes that have the smallest center distance up to a certain threshold. Then the performance (Recall) is the statistics for the matched part of all ground-truth boxes in the FOV of cameras.
The final results are averaged over the matching thresholds of $(0.5, 1, 2, 4)$ meters.
We can draw several conclusions:
\begin{enumerate}[1)]
    \item Over 50\% of objects (\textit{Vehicle} and \textit{Pedestrian}) are recalled in the 1m range compared to ground truths, showing the accuracy of our multi-modal spatial alignment and parallax occlusion and noise filtering.
    \item Almost 70\% to 80\% of objects are recalled in the 4m range, demonstrating vision foundation models possess sufficient capability to provide detection proposals, and the majority of objects not recalled are primarily due to occlusion (cameras are installed at a lower position relative to the LiDAR).
    \item There exists a distance gap between predictions and ground-truths (almost 20\% of objects are in the range of 1$\sim$4m), which is mainly due to the processing pipeline of our box interpretation. Previous box prediction of 3D detection models are always separately to predict the components, \eg, the CenterHead of CenterPoint predicts the box center, geometry size, and heading direction with different network layers. Different from them, our box interpretation would first predict the geometry size, and then calculate the center with the half of length, height, and width. Therefore, if the geometry size is inaccurate, the box center will also not be precise.
\end{enumerate}

\begin{table}[!h]
\small
\setlength{\tabcolsep}{0.32cm}
\caption{Detailed performance of 3D object detection on WOD val set. Metrics are Recall (with BEV distance criterion) of L2 difficulties for \textit{Vehicle}, \textit{Pedestrian}, and \textit{Cyclist}. All results are in the FOV of camera views.}
\label{tab:app-det-recall}
\begin{center}
\renewcommand{\arraystretch}{1.2}
  \begin{tabular}{l|ccccc}
    \Xhline{0.75pt}
    
    & Avg. & 0.5m & 1m & 2m & 4m \\

    \hline

    \textit{Vehicle} & $61.6$ & $39.2$ & $54.9$ & $70.7$ & $81.6$ \\

    \textit{Pedestrian} & $58.5$ & $42.5$ & $57.2$ & $64.5$ & $70.1$ \\

    \textit{Cyclist} & $34.0$ & $25.4$ & $32.8$ & $37.5$ & $40.4$ \\
    
    \Xhline{0.75pt}
  \end{tabular}
  \end{center}
\end{table}

\subsection{Performance of Occlusions}
\label{sec:app-occlusion}
We report the L1 AP performance of the overall and the occlusion part on WOD validation set to compare with other methods. The occlusion levels are defined based on whether the objects are obscured in the image perspective, which are provided by WOD.

As shown in Tab.~\ref{tab:app-det-occlusion}, compared to the overall performance, the occlusion part of CenerPoint, VoxelRCNN and PVRCNN exhibit decreases of 18.81\%, 18.78\% and 19.07\% respectively, while SAM3D shows a decrease of 31.30\%. In contrast, our method demonstrates a decrease of only 11.02\%. This improvement is attributed to our mask tracking module, which effectively leverages temporal context to mitigate the influence of occlusion.

\begin{table}[!h]
\small
\setlength{\tabcolsep}{0.32cm}
\caption{Performance comparison of occlusion on WOD val set. Metrics are L1 AP (with IoU criterion) for \textit{Vehicle}. All results are in the FOV of camera views.}
\label{tab:app-det-occlusion}
\begin{center}
\renewcommand{\arraystretch}{1.2}
  \begin{tabular}{l|ccc}
    \Xhline{0.75pt}
    
    & Training Data & Overall & Occluded \\

    \hline

    CenterPoint & \textit{train set} & $73.04$ & $59.30$ \\

    VoxelRCNN & \textit{train set} & $76.29$ & $61.96$ \\

    PVRCNN & \textit{train set} & $75.53$ & $61.13$\\

    SAM3D & - & $6.90$ & $4.74$ \\

    ZOPP (ours) & - & $37.56$ & $33.42$ \\
    
    \Xhline{0.75pt}
  \end{tabular}
  \end{center}
\end{table}

\subsection{Parallax Occlusion and Noise Filtering}
\label{sec:app-parallax}

\begin{figure*}[!h]
\centering
\includegraphics[width=1.\textwidth]{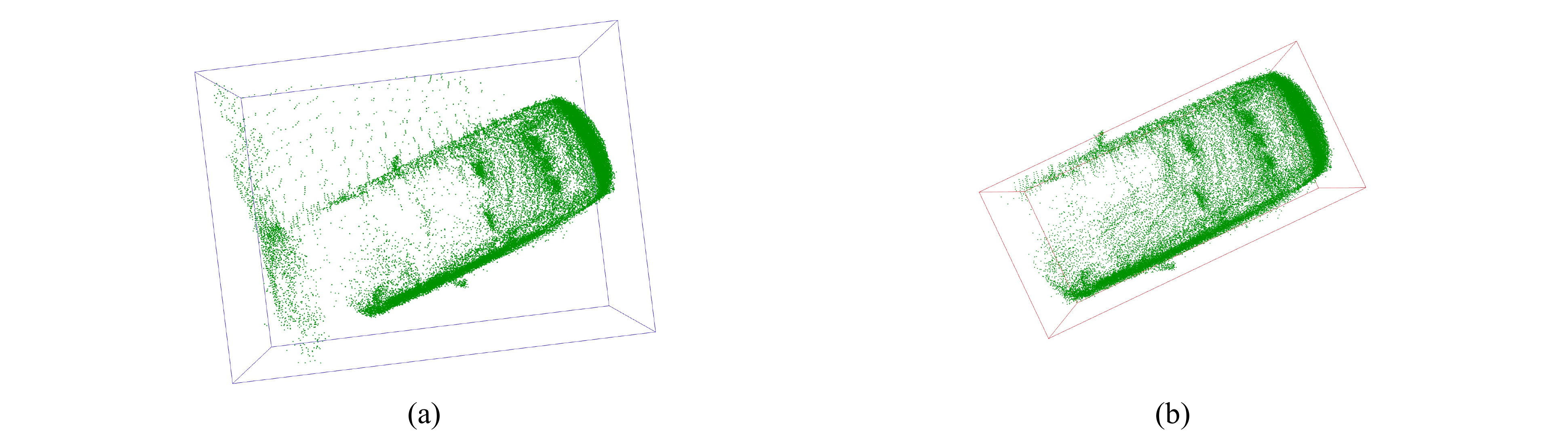}
\caption{(a) Before the parallax occlusion and noise filtering, our box interpretation would produce inaccurate box dimensions based on the wrong object points. (b) After filtering, we will output 3D boxes with precise dimensions.}
\label{fig:parallax-box}
\end{figure*}

We present the effectiveness of our parallax occlusion and noise filtering module by comparing the box interpretation results before and after the filtering operation.
As shown in Fig.~\ref{fig:parallax-box}, if we assign the instance and category information to the points that are directly projected to the image plane, some background points would be classified as the foreground objects, resulting in incorrect box interpretation. Our method could filter out the background and noise points, which significantly reduces the burden of our box interpretation module.

\begin{figure*}[!h]
\centering
\includegraphics[width=1.\textwidth]{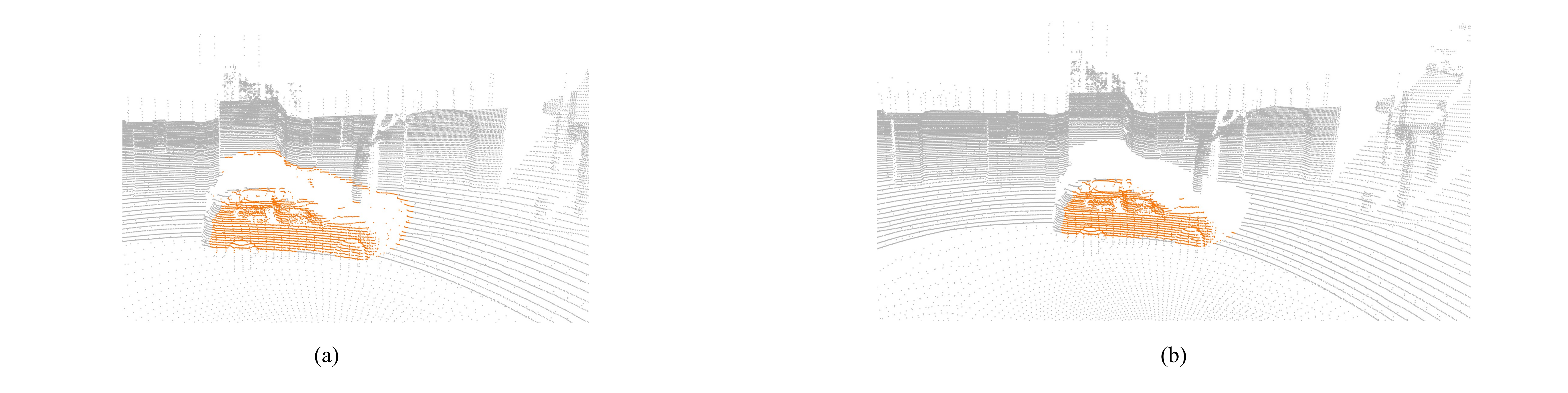}
\caption{(a) Before the parallax occlusion and noise filtering, we would assign the instance ID or semantic category of foreground objects to background points. (b) After the filtering operation, the segmentation results would be more accurate.}
\label{fig:parallax-seg}
\end{figure*}

In addition, we evaluate its effect on the segmentation task. We first report the performance of semantic segmentation in Tab.~\ref{tab:app-seg-abl}, which shows a significant quantitative improvement for all foreground objects and backgrounds. For the objects that always appear at higher altitudes (\eg, sign, traffic light), the parallax occlusion issue is not serious, hence the performance is maintained the same after the filtering module.
We also show the visualization comparison of segmentation results in Fig.~\ref{fig:parallax-seg}. As we can see, the background points may be located in the boundary regions of the foreground car, hence the corresponding categories are all incorrectly assigned as car. Our filtering module can filter out these points and better align the relation between LiDAR points and image pixels, producing more accurate segmentation results.

\begin{table*}[!h]
\caption{Comparisons of ZOPP on semantic segmentation before and after the proposed parallax occlusion and noise filtering.}
\label{tab:app-seg-abl}
\setlength{\tabcolsep}{0.004\linewidth}
\centering
\begin{adjustbox}{width=0.99\columnwidth,center}
\begin{tabular}{l| c  c c c c c c c c c c c c c c c c c c}
    \toprule
    Method 
    & \rotatebox{90}{Vehicle} 
    & \rotatebox{90}{motorcyclist}
    & \rotatebox{90}{bicyclist} 
    & \rotatebox{90}{pedestrian} 
    & \rotatebox{90}{sign} 
    & \rotatebox{90}{traffic light} 
    & \rotatebox{90}{pole} 
    & \rotatebox{90}{Cons. Cone} 
    & \rotatebox{90}{bicycle} 
    & \rotatebox{90}{motorcycle} 
    & \rotatebox{90}{building} 
    & \rotatebox{90}{vegetation} 
    & \rotatebox{90}{tree trunk}
    & \rotatebox{90}{curb}
    & \rotatebox{90}{road} 
    & \rotatebox{90}{lane marker}
    & \rotatebox{90}{other ground}
    & \rotatebox{90}{walkable}
    & \rotatebox{90}{sidewalk} \\
    \midrule

    ZOPP$^\dagger$ (before) & 51.6 & - & 47.7 & 76.1 & 29.5 & 34.0 & 49.6 & 32.7 & 21.2 & 34.2 & 73.8 & 72.3 & - & - & 80.5 & - & - & - & 60.4 \\

    ZOPP$^\dagger$ (after) & 54.2 & - & 49.6 & 77.3 & 29.7 & 34.2 & 51.7 & 33.1 & 21.8 & 35.4 & 75.5 & 73.6 & - & - & 81.8 & - & - & - & 61.2 \\

\bottomrule
\end{tabular}
\end{adjustbox}
\end{table*}

\subsection{Point completion}
\label{sec:app-exp-pcc}
\begin{figure*}[!h]
\centering
\includegraphics[width=1.\textwidth]{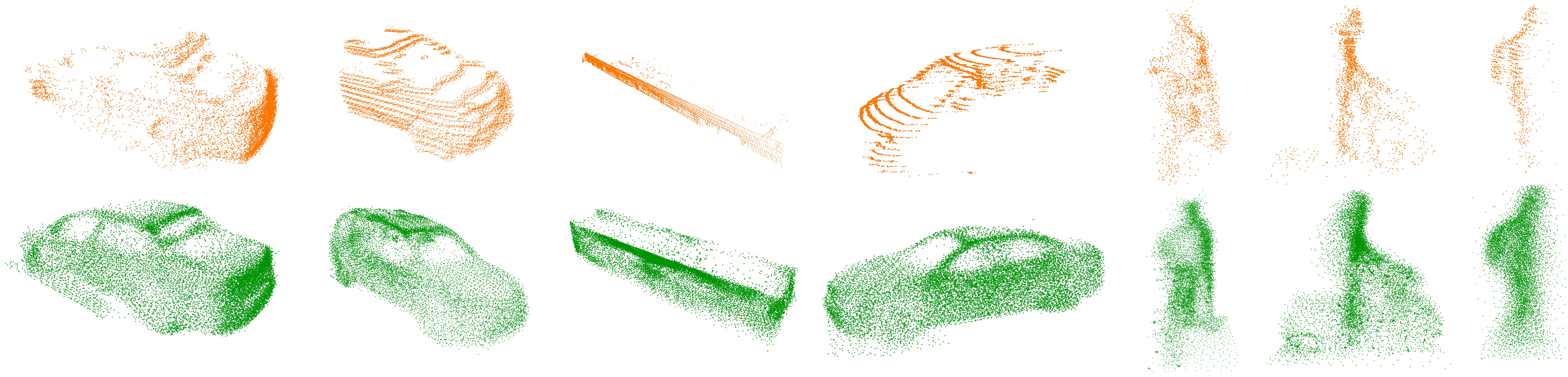}
\caption{Visual comparisons of point cloud completion. Compared with the sparse inputs (\textbf{\textcolor{myorange}{Top}}), we can produce fine-grained geometric structures of dense point clouds (\textbf{\textcolor{mygreen}{Bottom}}).}
\label{fig:vis-pcc}
\end{figure*}

We first evaluate the performance of point completion, we visualize the generated dense and completed object points shown in Fig.~\ref{fig:vis-pcc}.
The input object points are always sparse (1st, 5th columns) and uncompleted (2nd, 3rd, 4th columns), \eg, the bus in the 3rd column only has points at the top of the side surface. As a comparison, the generated dense points contain much more geometric structures, which would contribute much to interpreting precise 3D bounding boxes.
The high-quality results can also be used for generative assets modeling in simulation applications.

\begin{figure*}[!h]
\centering
\includegraphics[width=0.95\textwidth]{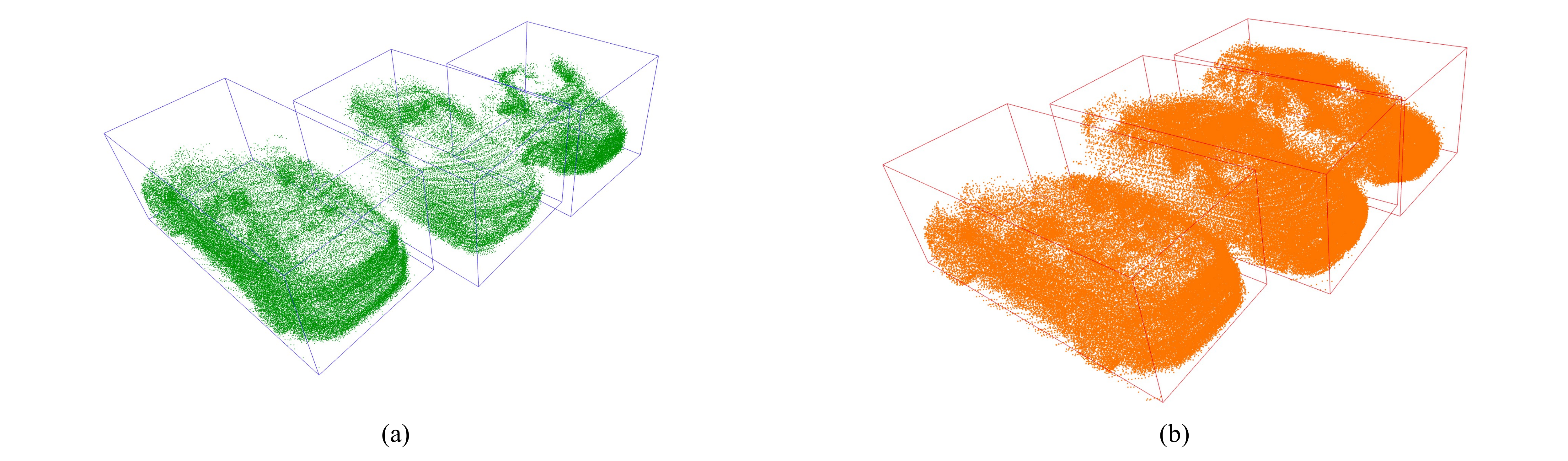}
\caption{(a) The object points are always sparse and partial, which would lead to inaccurate box interpretation. (b) After point completion for each object, we will predict 3D boxes with precise dimensions.}
\label{fig:pcc-box}
\end{figure*}

Afterward, to better evaluate the effectiveness of point completion in our pipeline, we visualize the interpreted 3D bounding boxes based on the object points processed before and after the point completion in Fig.~\ref{fig:pcc-box}. It is crucial to first generate complete point clouds for our human-label-free box interpretation module, resulting in accurate geometric sizes (length, width, height) prediction.

Furthermore, we report the quantitative results that would reflect the improvements of both the filtering and completion modules.
We calculate the Recall performance based on IoU criterion, which considers the accuracy of box shapes.
As shown in Tab.~\ref{tab:app-abl-pcc}, after the point completion process, the Recall is gained with $26.2$, $12.5$, and $11.4$ points on the three categories.
Because vehicles are always larger than the other two categories, it is more likely to produce sparse and incomplete point clouds. So, our completion module shows an impressive effect for our 3D box interpretation module.

\begin{table}[!h]
\setlength{\tabcolsep}{0.32cm}
\caption{Verifying the effect of parallax noise filtering and point completion for 3D bounding box interpretation on WOD val set. Metrics are Recall of L2 difficulties for \textit{Vehicle}, \textit{Pedestrian}, and \textit{Cyclist} with IoU criterion. The results are in the FOV of the cameras.}
\label{tab:app-abl-pcc}
\begin{center}
\renewcommand{\arraystretch}{1.2}
  \begin{tabular}{l|ccc}
    \Xhline{0.75pt}
     & \textit{Vehicle} & \textit{Pedestrian} & \textit{Cyclist} \\
     \hline

    Before & {$22.6$} & {$34.2$} & {$11.5$} \\
    
    After & {$48.8$} & {$46.7$} & {$22.9$} \\
    
    \Xhline{0.75pt}
  \end{tabular}
  \end{center}
\end{table}

\subsection{Open-set 3D Detection}
\label{sec:app-openset}
\begin{figure*}[h]
\centering
\includegraphics[width=1.\textwidth]{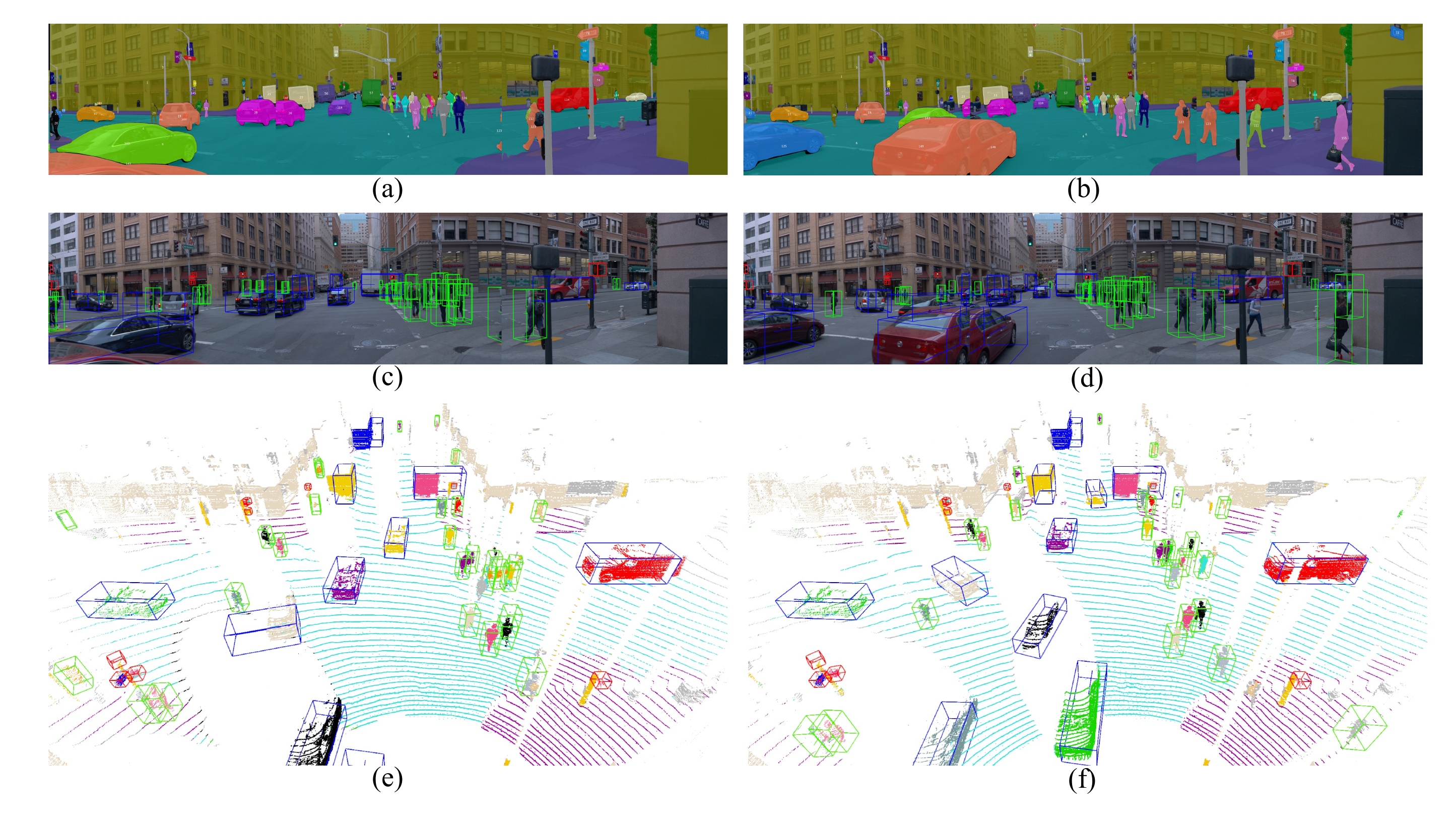}
\caption{The open-set detection results of ZOPP on WOD in both 2D and 3D spaces for consecutive frames.}
\label{fig:good_case}
\end{figure*}

As shown in Fig.~\ref{fig:good_case}, our ZOPP can output the open-set 3D detection results of traffic sign and traffic light (represented with red color bounding boxes). However, for elevated traffic lights beyond LiDAR scanning range, our method cannot generate corresponding 3D boxes due to the absence of point cloud data.

\subsection{Failure Pattern Analysis}
\label{sec:app-failure}
\begin{figure}[ht]
\centering
\includegraphics[width=1.\textwidth]{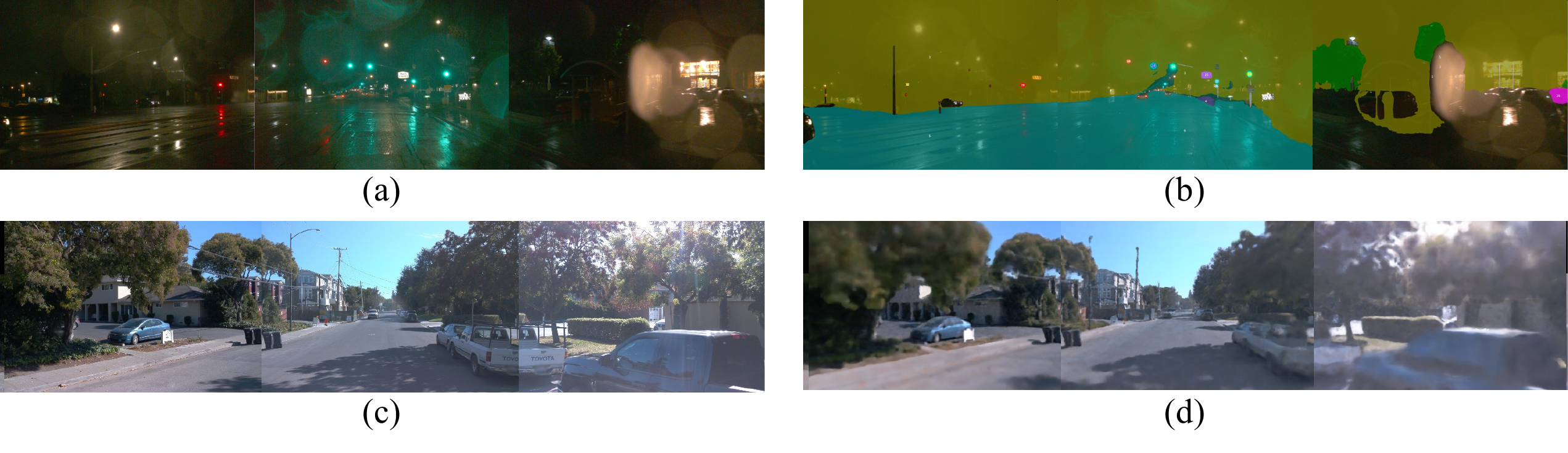}
\caption{The illustration of the failure cases. It indicates that the image data are influenced by the lighting conditions at night (a), rainy weather conditions (a), and the camera's overexposure condition (c). Then we could not generate accurate detection and segmentation results (b), and reconstruction with lower quality (d).}
\label{fig:bad_case}
\end{figure}

We have briefly summarized some representative challenging scenarios in Sec.~\ref{sec:limit} of our main contents. Firstly, our method would fail to effectively recognize similar object categories (e.g., construction vehicle, truck, trailer) and some uncommon object categories (e.g., tree trunk, lane marker) with the foundation models (Grounding-DINO). Since this is the first stage of our entire method, it will result in subsequent stages lacking the output of corresponding perception results, such as 3D segmentation and occupancy prediction. Secondly, neural rendering methods may encounter numerous challenges in street-view scenes, constrained by practice factors (adverse weather conditions, sensor imaging issues), such as camera overexposure. Our occupancy decoding will fail in these scenarios where it is impossible to generate geometrically plausible 3D reconstructions. Please refer to Fig.~\ref{fig:bad_case} for qualitative visualizations.

\clearpage
\newpage

\FloatBarrier

\clearpage

\end{document}